\documentclass[letterpaper]{article} 
\usepackage{aaai2026}  
\usepackage{times}  
\usepackage{helvet}  
\usepackage{courier}  
\usepackage[hyphens]{url}  
\usepackage{graphicx} 
\usepackage{natbib}  
\usepackage{caption} 
\usepackage{algorithm}
\usepackage{algorithmic}

\usepackage[T1]{fontenc}
\usepackage{subfigure}
\usepackage{makecell}
\usepackage{threeparttable}
\usepackage{multirow}
\usepackage{amsthm,amsmath,amssymb}
\usepackage{mathrsfs}
 \usepackage{cite} 
\usepackage{mathptmx}
\usepackage{amssymb}
\usepackage{bbding}

\usepackage{newfloat}
\usepackage{listings}
\DeclareCaptionStyle{ruled}{labelfont=normalfont,labelsep=colon,strut=off} 
\floatstyle{ruled}
\newfloat{listing}{tb}{lst}{}
\floatname{listing}{Listing}
\title{MindSpeed RL:  Distributed Dataflow for Scalable and Efficient RL Training on Ascend NPU Cluster}
\author{
    Liangjun Feng \textsuperscript{\rm 1}
    Chenyi Pan ,
    Xinjie Guo ,
    Fei Mei ,
    Benzhe Ning ,
    Jianxiang Zhang ,
    Xinyang Liu,
    Beirong Zhou ,
    Zeng Shu ,
    Chang Liu ,
    Guang Yang ,
    Zhenyu Han ,
        Jiangben Wang  ,
     Bo Wang  
}
\affiliations{
    \textsuperscript{\rm 1} Huawei Lianqiu Lake R\&D Center \\

    Xicen Community, Jinze Town, Qingpu District\\
    Shanghai 201718 China\\
    fengliangjun@huawei.com
}

\usepackage{bibentry}

\begin{document}

\maketitle

\begin{abstract}
Reinforcement learning (RL) is a paradigm increasingly used to align large language models. Popular RL algorithms utilize multiple workers and can be modeled as a graph, where each node is the status of a worker and each edge represents dataflow between nodes. Owing to the heavy cross-node dependencies, the RL training system usually suffers from poor cluster scalability and low memory utilization.  In this article, we introduce MindSpeed RL, an effective and efficient system for large-scale RL training. Unlike existing centralized methods, MindSpeed RL organizes the essential data dependencies in RL training, i.e., sample flow and resharding flow, from a distributed view.  On the one hand, a distributed transfer dock strategy, which sets controllers and warehouses on the basis of the conventional replay buffer,  is designed to release the dispatch overhead in the sample flow. A practical allgather--swap strategy is presented to eliminate redundant memory usage in resharding flow. In addition, MindSpeed RL further integrates numerous parallelization strategies and acceleration techniques for systematic optimization. Compared with existing state-of-the-art systems, comprehensive experiments on the RL training of popular Qwen2.5-Dense-7B/32B, Qwen3-MoE-30B, and DeepSeek-R1-MoE-671B show that MindSpeed RL increases the throughput by $1.42 \sim 3.97$ times. Finally, we open--source MindSpeed RL and perform all the experiments on a super pod of Ascend with 384 neural processing units (NPUs) to demonstrate the powerful performance and reliability of Ascend.
\end{abstract}

\begin{links}
    \link{Code}{https://gitee.com/ascend/MindSpeed-RL}
\end{links}

\section{Introduction}
Reinforcement learning (RL) has been validated to be effective in improving the reasoning and aligning capabilities of large language models (LLMs) \cite{openai2024gpt4technicalreport, wang2025reinforcementlearningenhancedllms}. State-of-the-art LLMs such as Qwen3 \cite{yang2025qwen3technicalreport}, DeepSeek-R1 \cite{deepseekai2025deepseekr1incentivizingreasoningcapability}, and OpenAI-o1 \cite{openaio1} all benefit from the inference--time scaling law of RL to achieve high scores on mathematics and coding tasks. However, RL training is much more complicated than supervised training. Mainstream RL algorithms, such as GRPO \cite{shao2024deepseekmathpushinglimitsmathematical} and PPO \cite{2017Proximal}, usually involve multiple workers, e.g., actor workers and reference workers, and switch states between training and generation. Owing to the scheduling and communication dependencies between workers and states, constructing an efficient large-scale RL training system is generally challenging in practice \cite{2021Colossal, aframework}. \par
\begin{figure}[htb]
\centering
\includegraphics[width=.47\textwidth, trim=1mm 1mm 1mm 1mm, clip]{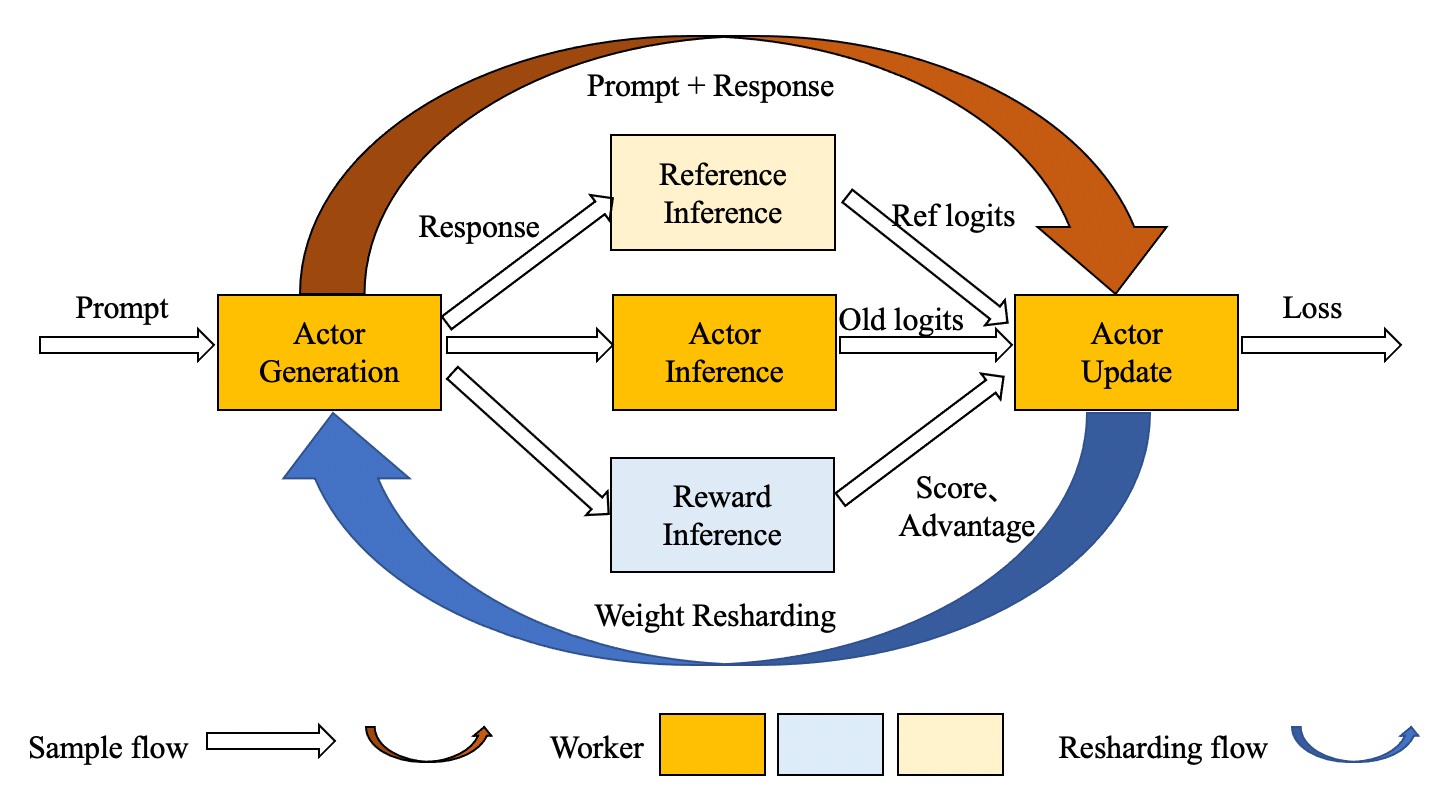}
\caption{Data dependencies in RL, including sample flow and resharding flow. GRPO is used here as an example of an RL algorithm, where workers are usually implemented on the basis of LLMs.}
\end{figure}
As depicted in Figure 1, data dependencies in RL can be divided into two types, i.e., sample flow and resharding flow \cite{2024HybridFlow}. Specifically, the sample flow describes how the algorithm is performed on the basis of a prompt input. The responses generated by  actor workers are sent with their prompts to reference workers and reward workers to obtain transient data, such as logits and reward scores  \cite{liang2021rllibflowdistributedreinforcement}.  Moreover, samples flow along different states of actor workers, i.e. generation, inference, and update states, to obtain the final loss \cite{2022Training}. The resharding flow transfers the updated actor weights from the training state to the generation state to adjust the parallelization strategies \cite{2024HybridFlow}. With increasing model size and cluster size of RL training, the RL system becomes increasingly complex, which challenges hardware efficiency \cite{trlx}. Acceleration devices, i.e., general processing units (GPUs) and neural processing units (NPUs), are expensive, and insufficient systems exacerbate this problem and sometimes fail at the task. A scalable and efficient framework for RL training is desirable; hence, effective LLMs with reasoning ability can be obtained with a reasonable budget.  \par

Owing to the importance of efficiency issues, several solutions have been proposed for RL. OpenRLHF \cite{2024OpenRLHF} designs model scheduling using Ray \cite{Philipp2017Ray}, vLLM \cite{zhang2025jengaeffectivememorymanagement}, and DeepSpeed \cite{rajbhandari2022deepspeedmoeadvancingmixtureofexpertsinference, 2020DeepSpeed}, which enables training of models with more than 70 billion parameters. ReaL \cite{mei2025realefficientrlhftraining} redistributes LLM parameters across the training cluster for different workloads to overcome the low efficiency of fixed parallelization strategies. Areal \cite{fu2025areallargescaleasynchronousreinforcement} focuses on the long-tail generation problem caused by synchronous mechanism, and uses a modified asynchronous objective of the PPO algorithm to improve the training throughput. The asynchronous strategy is also used in StreamRL \cite{zhong2025streamrlscalableheterogeneouselastic}, which overlaps the execution of the generation and inference stages. RLHFuse \cite{zhong2025optimizingrlhftraininglarge} applies interstage fusion and intrastage fusion to reduce the number of system bubbles and improve the hardware utilization. \par

However, in practical deployment, the performance bottleneck of RL training is not only the update or generation loads but also the complex data dependency and dataflow among workers, which sometimes take up more than half of the time and cause redundant memory usage. We further detail this challenge in the problem statement section. Few articles have explored the dataflow problem at scale. K1.5 \cite{kimiteam2025kimik15scalingreinforcement} intuitively constructs a replay buffer that acts as a central storage mechanism for RL. The centralized method generally struggles with increasing cluster size. VeRL \cite{2024HybridFlow} performs fine-grained weight resharding to reduce the transition overhead between training and generation, but it ignores the elimination of redundant memory usage among computing states. Basically, the optimization of an RL system is supposed to be comprehensive, where all of the efficiencies of computing, communication, and memory are needed. \par

To resolve the above issues in system design, we propose MindSpeed RL, a novel RL framework for superior training efficiency. First, a distributed transfer dock strategy is designed for sample flow, which splits controllers and warehouses from the conventional replay buffer. Each worker state, e.g., actor generation or reward inference, is assigned its own data controller, thereby alleviating congestion caused by cross-node requests. Moreover, multiple warehouses are set across the cluster to split the sample flow along the global batch size. The distributed controllers and warehouses collaborate to release the dispatch overhead in the sample flow. Second, we present a practical allgather--swap technique for resharding flow to eliminate redundant memory  usage in distributed training. Specifically, in resharding flow, we allgather weights for actor generation from the update state, and then move the redundant memory buffer of the update state to the host, thereby increasing the available memory for actor generation. During actor update, the memory buffer on the host is moved back to the device for the next round of training. On the basis of the high bandwidth between the host and device, the proposed allgthaer--swap technique eliminates all of the redundant memory usage in resharding flow with a negligible duration. \par

In addition, we integrate numerous parallelization strategies and acceleration techniques into MindSpeed RL to perform systematic optimization for the computing, communication, and memory of RL training. Comprehensive experiments on the popular Qwen2.5-Dense-7B/32B, Qwen3-MoE-30B, and DeepSeek-R1-MoE-671B are conducted on a super pod of Ascend to validate the powerful performance and reliability of MindSpeed RL and Ascend. We emphasize the value and contribution of open--sourcing these codes and data. To the best of the author's knowledge, most open--source frameworks cannot train all of these models.\par

Our contributions are summarized as follows:
\begin{itemize}
    \item We identify the dataflow bottlenecks in RL  including sample flow and resharding flow, which could cause dispatch overhead and redundant memory problems, respectively.
    \item We design a distributed dataflow mechanism, i.e., the transfer dock strategy and allgather--swap technique, to improve dispatch efficiency and memory utilization. 
    \item We develop an open--source MindSpeed RL framework, which integrates numerous parallelization strategies and acceleration techniques to support the scalable and efficient RL training of LLMs.
    \item We conduct comprehensive evaluations on a super pod of Ascend with 384 NPUs, where the maximum model size reaches 670 billion. Compared with the baselines, MindSpeed RL achieves $1.42 \sim 3.97$ times higher throughput.   
\end{itemize}

\section{Background}
\subsection{RL Workflow}
The workflow of mainstream RL algorithms, e.g., GRPO \cite{shao2024deepseekmathpushinglimitsmathematical}, PF-PPO \cite{yao2023deepspeedchateasyfastaffordable}, and DAPO \cite{yu2025dapoopensourcellmreinforcement}, includes the following stages:
\subsubsection{Generation stage.} Given prompts as input, the actor worker using LLM generates responses for RL as training data. To support large-scale LLMs, generation engines, e.g., vLLM, TensorRT \cite{tensorrt}, and SGLong \cite{2023arXiv231207104Z}, which have various parallelization strategies, are usually required by actor worker. Among these generation engines, vLLM is especially popular because of its user-friendly interface. Hence, we use vLLM--Ascend \cite{vllmascend} in MindSpeed RL, which is a plugin for vLLM on the Ascend NPU.
\subsubsection{Inference stage.} Most of the workers, e.g., reference worker and reward worker, would perform a forward pass on the generated samples. The output logits are used to calculate the training loss in the update stage \cite{noukhovitch2025asynchronousrlhffasterefficient, shen2025exploringdatascalingtrends, 2022Chain}.  
\subsubsection{Update stage.} The actor worker updates the model parameters on the basis of the results from the generation and inference stages. Hence, a training engine is also required by the actor worker. Popular training engines for LLM include DeepSpeed, FSDP2 \cite{fsdp2, 2023PyTorch}, Megatron-LM \cite{2021Efficient}, and MindSpeed \cite{mindspeed}. Here, MindSpeed RL uses MindSpeed as its training engine, which was developed on the basis of Ascend. After the actor worker completes the parameter update, it switches to the generation state and prepares for the generation in the next iteration.
\subsubsection{Data scheduler.} As mentioned above, an RL system uses sample flow and resharding flow to connect the generation stage, inference stage, and update stage. The scheduler in existing frameworks is either missing \cite{2024HybridFlow, mei2025realefficientrlhftraining, 2024OpenRLHF} or intuitively implemented as a memory buffer, namely a replay buffer \cite{kimiteam2025kimik15scalingreinforcement, fu2025areallargescaleasynchronousreinforcement}. The naive centralized method presents inferior performance in scalable cases. In this paper, we distribute the dataflow to address the bottleneck.

\subsection{Acceleration Techniques}
To scale up RL training for LLMs, several useful techniques have been proposed. Here, we introduce these techniques to demonstrate the superiority of MindSpeed RL.
\subsubsection{Parallelization strategies.} For dense models, data parallelism (DP) \cite{2019Megatron, 2021Efficient}, tensor parallelism (TP) \cite{ 2022Accelerating, singh2025modelparallelismsubnetworkdata}, and pipeline parallelism (PP) \cite{2020GPipe, 2023Zero} are commonly used. DP replicates the model weights across devices and splits the input data among them. TP divides individual operations, e.g., the matmul operation, across multiple devices. PP splits the transformer layers into separate stages, each assigned to a different device. DP, TP and PP alleviate the computing and memory pressure on individual devices at the cost of communication. For mixture-of-experts (MoE) models, expert parallelism (EP) is designed, which distributes experts among devices \cite{rajbhandari2022deepspeedmoeadvancingmixtureofexpertsinference}. Because of the efficient all-to-all communication, EP is more practical for MoE models than TP and PP are. Additionally, for cases with long sequences, context parallelism (CP) is used to split the sequence into several parts and alleviate the memory pressure on attention computation \cite{0System, 2023Ring}. To summarize, DP, TP, PP, EP, and CP are essential techniques used to support the RL training of large-scale LLMs. MindSpeed RL comprehensively implements these parallelization strategies and their variants, e.g., ZeRO \cite{2020ZeRO}, VPP \cite{2019Megatron}, RingAttention \cite{2023Ring}, and Ulysses \cite{0System}, to present a practical and leading RL framework.
 \subsubsection{Fused kernels.} Merging multiple individual operations as a fused kernel is popular for the optimization of computing and memory. In addition to the famous FlashAttention kernel \cite{dao2023flashattention2fasterattentionbetter},  other useful kernels, e.g., RMSNorm \cite{graef2025flashnormfastnormalizationllms}, SwiGLU \cite{shazeer2020gluvariantsimprovetransformer}, RoPE \cite{su2023roformerenhancedtransformerrotary}, MatmulAdd \cite{2019Megatron}, and GMM \cite{mindspeed}, have been recognized for LLMs in recent years. On the basis of Ascend NPUs, we have implemented and integrated all of these fused kernels in MindSpeed RL for superior performance.
\begin{figure}[htb]
\centering
\includegraphics[width=.47\textwidth, trim=1mm 1mm 1mm 1mm, clip]{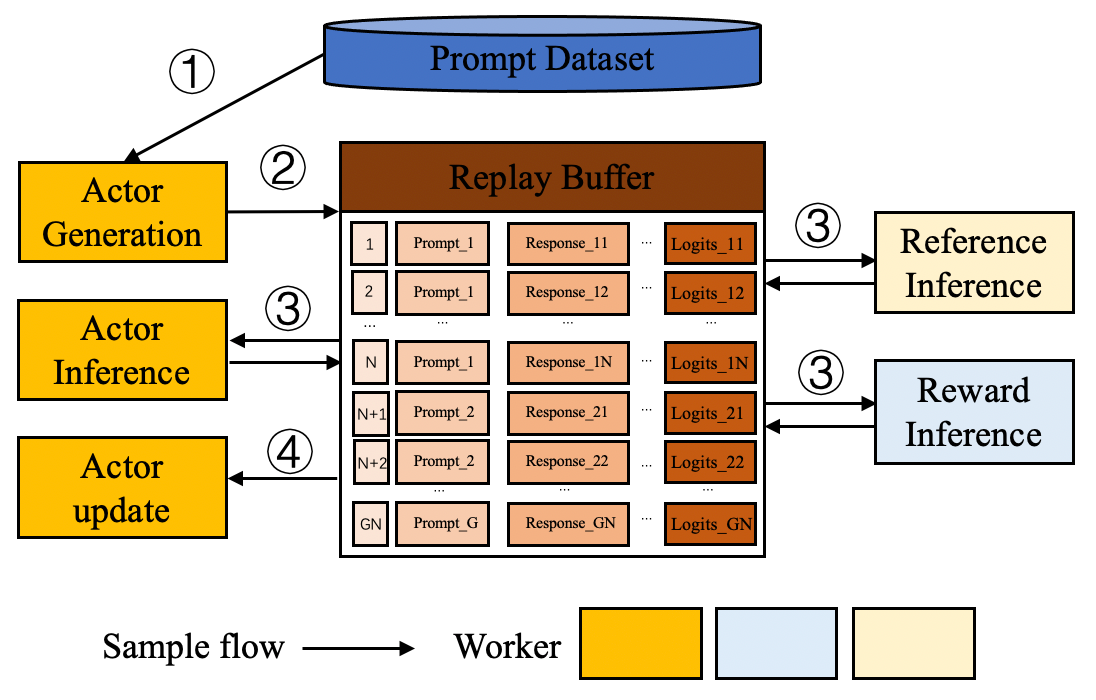}
\caption{Centralized replay buffer mechanism in RL systems, where $G$ is the global batch size of the input prompts and where $N$ is the number of responses generated for each prompt.}
\end{figure}

\begin{table}[!htb]
\centering
    \setlength{\tabcolsep}{2 mm}
    \centering
    \begin{tabular}{cccccc|c|cc}
        \Xhline{1pt}
        $G$ & $N$ & $PL$ & $n$ & $SL$ & $M$ & $TCV$ & $T100$ & $T1K$ \\
        \hline
        256  & 8  &  2K & 5  & 8K & 3 & 0.96 & 9.92 & 0.97  \\ 
        256  & 16  &  2K & 5  & 16K & 3 & 3.81 & 39.0 & 3.81 \\
        1K  & 16  &  2K & 5  & 16K & 3 & 15.2 & 156.1 & 15.2  \\ 
        1K  & 32  &  4K & 8  & 32K & 5 &97.0 & 993.3 & 97.0 \\
        4K  & 32 &  4K & 8  & 32K & 5 & 388.0 & 3.9K & 388.0  \\ 
        8K  & 64  &  4K & 8  & 64K & 5 & 3.1K & 31K & 3.1K\\
        \Xhline{1pt}
        \end{tabular}
            \textbf{\caption{TCV (GB) and dispatching time (s) with different configurations in RL systems, where $B$ in Eq. (2) is 4 for int32 or float32, and $T100$ and $T1K$ denote the dispatching times with 100 MB/s and 1 GB/s connections among servers, respectively. }}
\end{table}

\section{Methodology}
Here, we first demonstrate the dataflow problem and then introduce the two designed techniques and MindSpeed RL. 
\subsection{Problem Statement}
During practical deployment, two key issues were identified in the sample flow and resharding flow of existing RL systems, causing substantial device underutilization. \par
For sample flow, the replay buffer mechanism shown in Figure 2 is usually integrated, and works as a centralized data queue or storage mechanism  in RL systems. Compared with the naive sample flow shown in Figure 1,  the replay buffer enables the practical overlap techniques mentioned above, e.g., interstage fusion or partial rollout, to increase training efficiency. However, neither the centralized replay buffer mechanism nor the naive strategy addresses the dispatch overhead problem of sample flow. \par
For example, the actor worker in step 4 of Figure 2 requests a batch of data from the centralized replay buffer for parameter update, and the communication volume $CV$ (GB) in the GRPO algorithm can be estimated as follows:
 \begin{equation}
CV=G \times N \times B \times  (PL+n \times SL+M)/1024^{3},
\end{equation}
where $G$ is the global batch size, $N$ is the number of responses generated for each prompt, $B$ is the number of bytes of the data type, $PL$ is the maximum prompt length, $SL$ is the maximum response length, and $M$ is the number of scalars, such as index and response length. The number of items that have the same length as the responses, e.g., old logits and reference logits, is denoted $n$. With the last 3 steps in Figure 2, the total communication volume $TCV$ (GB) is as follows:
 \begin{equation}
TCV=G \times N \times B \times  ( 2PL+3n \times SL+8M ) /1024^{3}.
\end{equation}
On the basis of the above equation, the total volume and dispatching times with different hyperparameters are listed in Table 1. As shown, an RL system only spends a few seconds on sample flow with low loads. However, for scalable training in clusters with large hyperparameters, the sample flow costs thousands of seconds in an iteration. In practical deployment, the serialization and deserialization of Ray for tensors cost extra time and increase the dispatch time. In this study, we design a distributed transfer dock strategy to address the serious dispatch overhead problem. \par

\begin{figure}[htb]
\centering
\includegraphics[width=.47\textwidth, trim=1mm 1mm 1mm 1mm, clip]{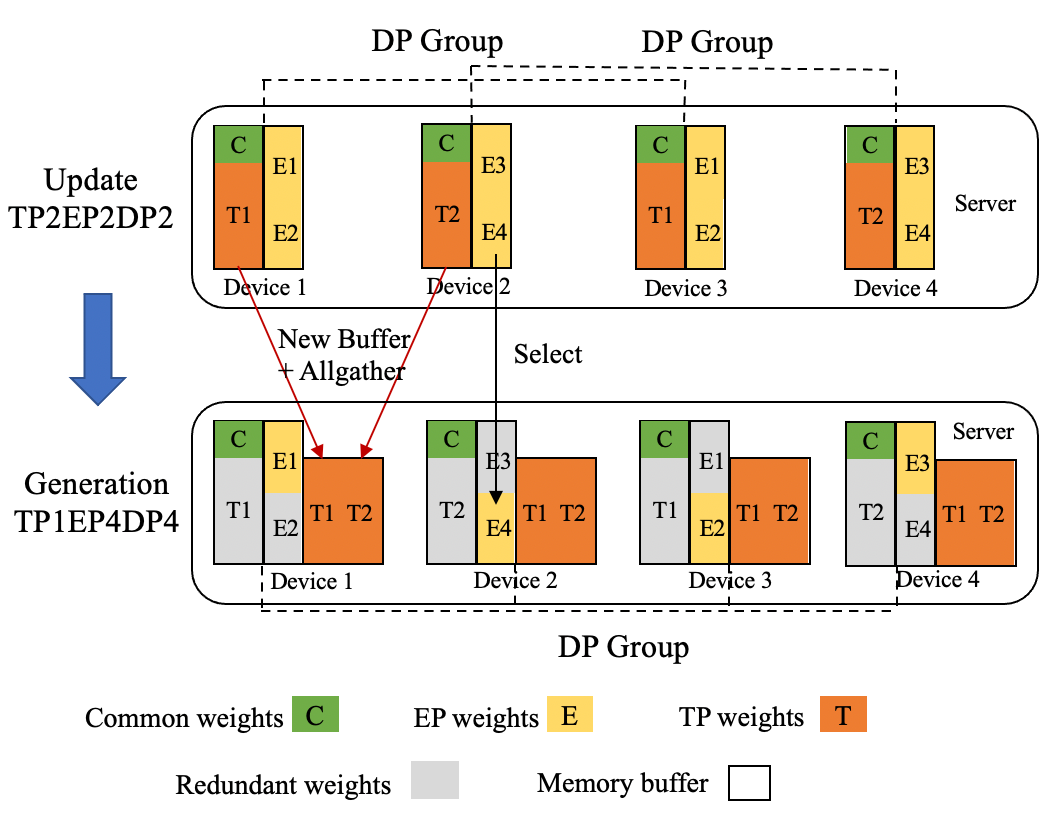}
\caption{Redundant memory issues in the distributed resharding flow of RL systems with 4 devices. }
\end{figure}

For resharding flow, we focus on redundant memory issues. A simple case where the update weights with TP2EP2DP2 are transferred to the generation weights with TP1EP4DP4 is shown in Figure 3. There are two redundant memory issues. First, before the allgather operation, a new buffer is applied for the TP weights, i.e., $T1$ and $T2$. Since $T1$ is placed in the same buffer with the common weights $C$, the original buffer cannot be released, resulting in redundant memory usage. Second, the EP weights $E3$ and $E4$ share a common buffer. In the generation stage, $E4$ can be directly selected for generation, but $E3$ is useless, leading to redundancy. The redundant memory $R$ (GB) in Figure 3 is estimated as follows:
 \begin{equation}
R = GDP \times (TW/UTP + EW/GEP), 
\end{equation}
where $GDP$ and $GEP$ are the DP size and EP size in the generation stage, respectively. The TP size in the update stage is denoted $UTP$. The sizes of the TP and EP weights are denoted $TW$ and $EW$, respectively.  For the Qwen3-MoE-30B model, the redundant memory in the resharding flow is more than 60GB. We address this problem with a new allgather--swap technique.

\begin{figure}[htb]
\centering
\includegraphics[width=.47\textwidth, trim=1mm 1mm 1mm 1mm, clip]{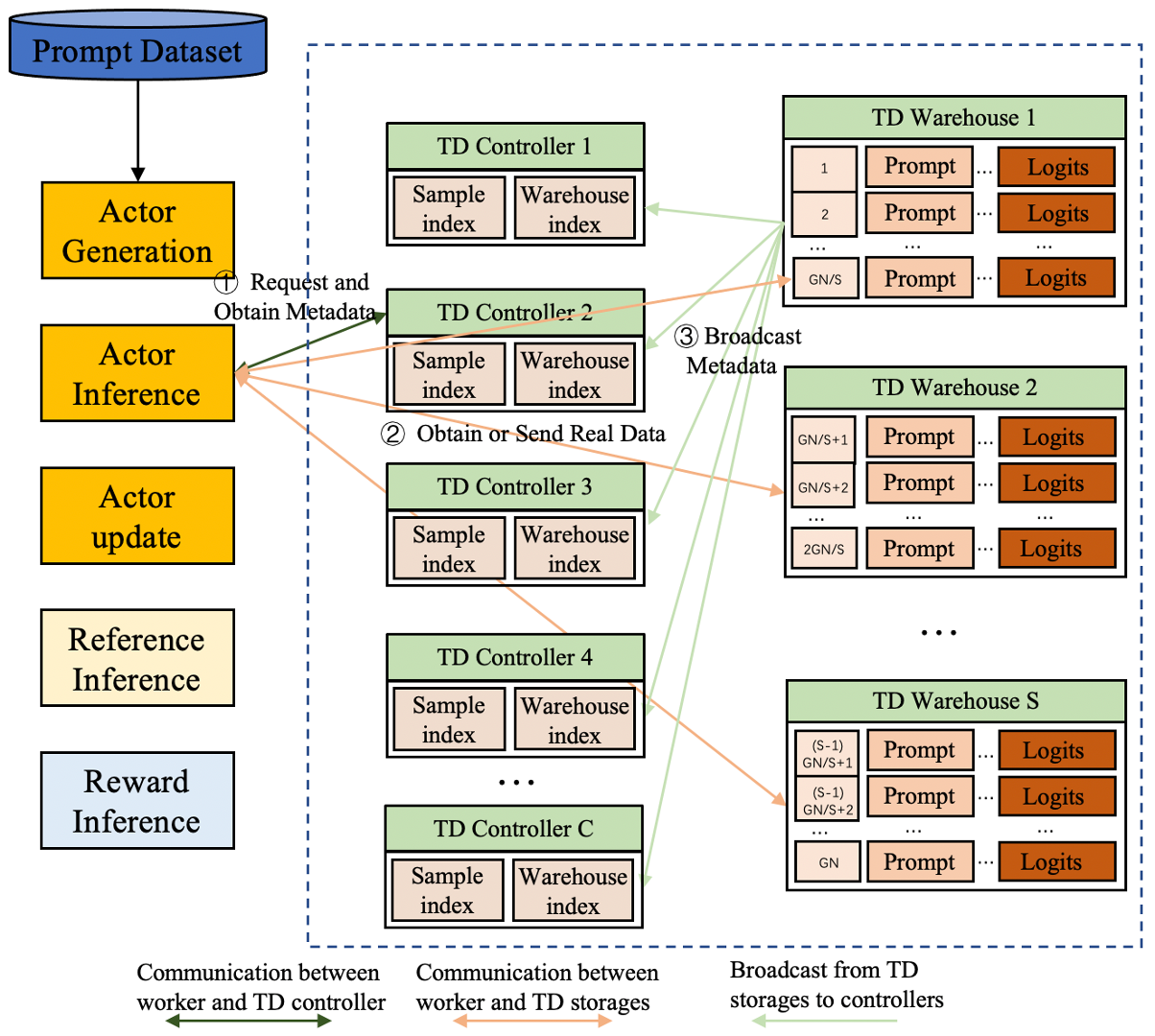}
\caption{Distributed transfer dock strategy. Each worker state has a TD controller, and the number of TD controllers is denoted $C$. The number of TD warehouses $S$ is usually equal to the number of nodes in the cluster. }
\end{figure}

\subsection{Distributed Transfer Dock Strategy}
The distributed transfer dock (TD) strategy sets controllers and warehouses on the basis of the conventional replay buffer, as shown in Figure 4. First, the centralized replay buffer is divided into multiple TD warehouses along the global batch size, where each sample has its own index. The number of TD warehouses $S$ is usually equal to the number of nodes in the cluster, and these warehouses are implemented in different nodes to share the dispatch loads. Second, a TD controller is assigned to each worker state, which records the metadata of the sample flow. The metadata, e.g., sample index and its warehouse index, enable the workers to obtain the required real data from distributed storage. The number of TD controllers is denoted $C$. The advantage of using multiple controllers is that we can integrate the controller with its worker together, thereby reducing the number of internode requests.\par
The workflow of the distributed transfer dock strategy in Figure 4 can be summarized in three steps. First, a worker state, such as the actor inference, requests and obtains metadata from its TD controller. Second, on the basis of metadata, the actor obtains real data from distributed storage. Note that a worker can also send real data to warehouses in this step. Finally, the TD warehouse updates its state and broadcasts the metadata to all controllers. Hence, on the basis of Eq. (2), the total communication volume for each TD warehouse is as follows:
 \begin{equation}
 \begin{aligned}
& TCV  = \\
& G \times N \times B \times  \left [  2PL+3n \times SL+8(C+1)M \right ] / S / 1024^{3}, 
\end{aligned}
\end{equation} 
where $C$ is the number of TD controllers and is usually $\left [5, 10 \right]$, which is decided by the specific RL algorithm. $S$ can be set as the cluster size, such as 16 or 128. Since the metadata are simple scalars, the $M$ in Eq. (4) is usually $\left [3, 5\right]$ as shown in Table 1; hence, the TD strategy in Eq. (4) can effectively address the dispatch overhead problem in RL training. In addition, we use TensorDict as the data structure of sample flow in transfer dock, which accelerates the serialization and deserialization processes of Ray in implementation. 

\begin{figure}[htb]
\centering
\includegraphics[width=.47\textwidth, trim=0.8mm 0.8mm 0.8mm 0.8mm, clip]{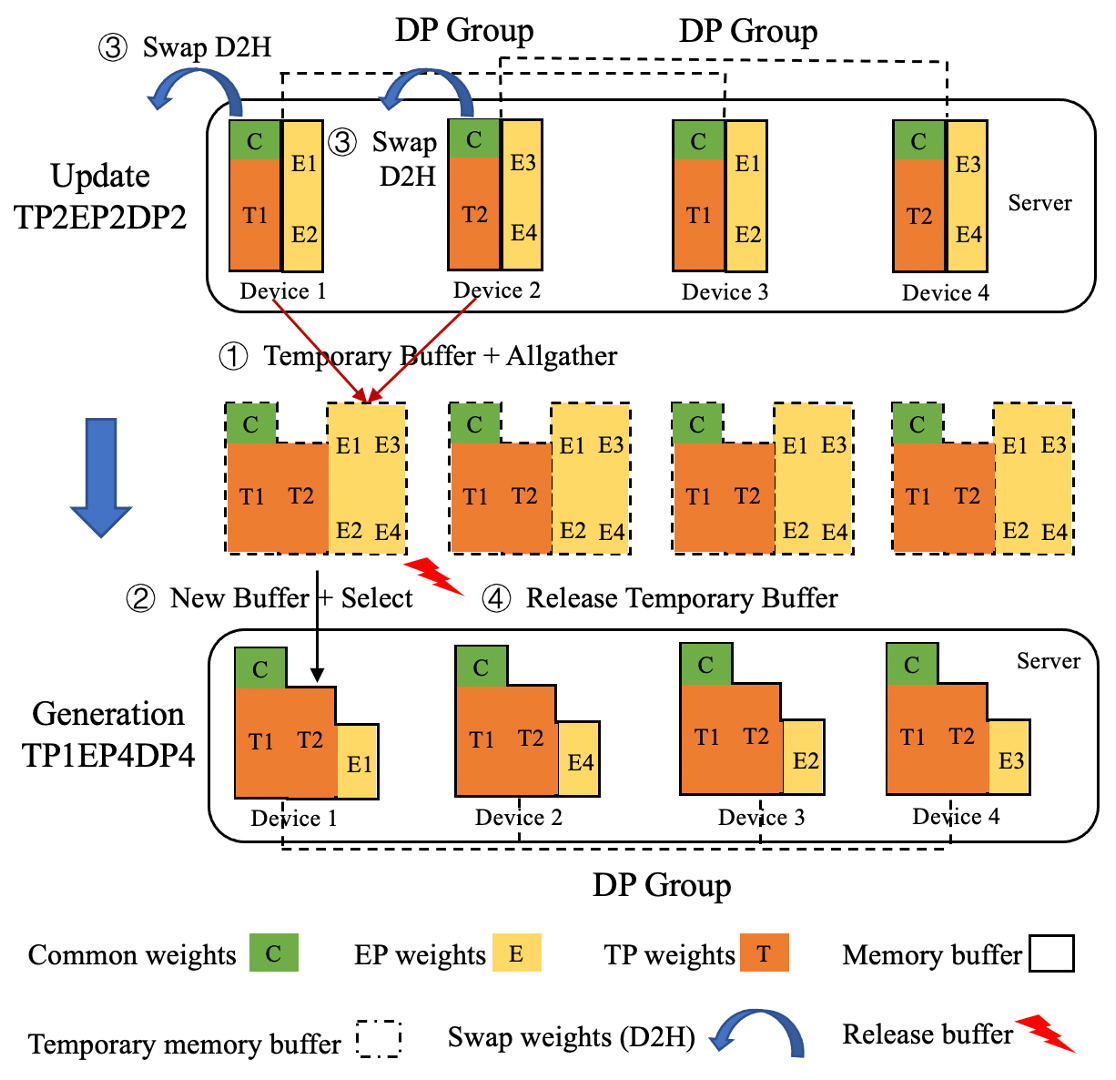}
\caption{Allgather--swap technique in the distributed resharding flow of RL systems with 4 devices, which addresses the redundant memory issues in Figure 3. }
\end{figure}
\subsection{Allgather--Swap Technique}
The allgather--swap technique addresses the redundant memory issues in the distributed resharding flow, which has four steps and is shown in Figure 5. First, a temporary buffer is applied for the allgather operation for model weights. Second, on the basis of the model weights in the temporary buffer, each device can select and copy the required weight slices for generation. Third, the original update weights are swapped from the device memory to the host memory (D2H). This step is essential for memory efficiency since it completely releases the memory buffer of update weights in the generation stage. Moreover, before the update stage of the next iteration, the update weights can be directly swapped from the host memory to the device memory (H2D) instead of performing a resharding operation again. Finally, the temporary buffer is released to finish the resharding flow. \par
Compared with the naive resharding method shown in Figure 3, the proposed allgather--swap technique introduces the swap operation between the host and device. Unlike the bandwidth between hosts shown in Table 1, the bandwidth between the host and device is usually greater than 50 GB/s, and the swap operation for model weights can be completed in a few seconds. Moreover, the H2D swap for the update stage can be performed in advance and overlapped with the inference stage for efficiency.

\begin{figure}[htb]
\centering
\includegraphics[width=.47\textwidth, trim=0.1mm 0.1mm 0.1mm 0.1mm, clip]{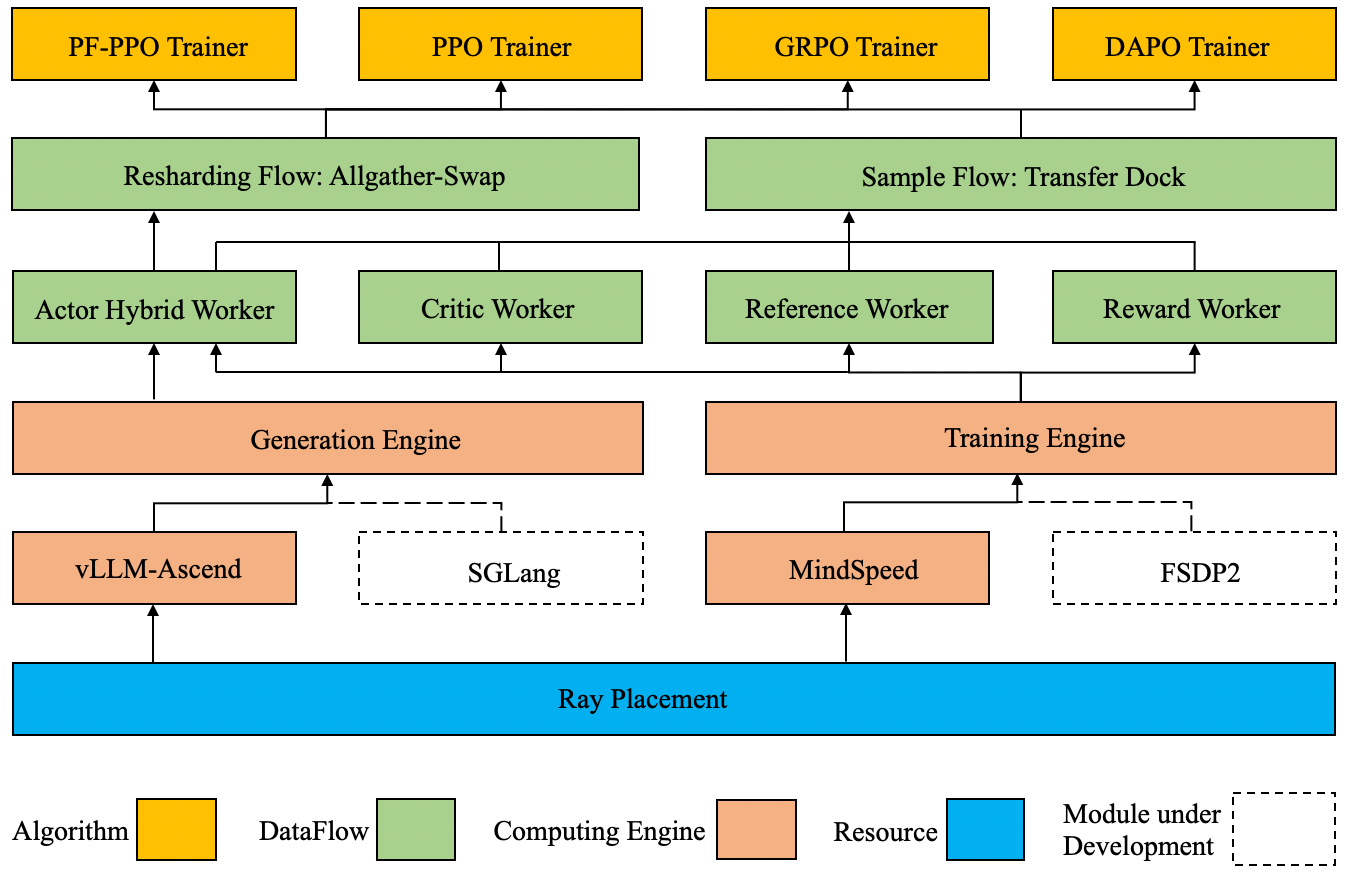}
\caption{Architecture of MindSpeed RL, which consists of four layers, i.e., a resource pool, a computing engine, a dataflow, and an algorithm.}
\end{figure}

\begin{table}[!htb]
\centering
    \setlength{\tabcolsep}{2 mm}
    \centering
    \begin{tabular}{c|cccc}
        \Xhline{1pt}
        Features & MSRL & OpenRLHF & VeRL & Trl  \\
         \hline
        PPO  & \checkmark  & \checkmark & \checkmark  &  \checkmark  \\
        GRPO  & \checkmark  &  \checkmark & \checkmark  &  \checkmark \\
        PF-PPO  & \checkmark  &   \XSolidBrush & \checkmark  &   \XSolidBrush \\
        DAPO  & \checkmark  &  \checkmark & \checkmark  &   \XSolidBrush \\
        DPO  & \checkmark  &  \checkmark &  \XSolidBrush  &  \checkmark  \\
        \hline
        DP  & \checkmark  &  \checkmark & \checkmark  &  \checkmark   \\ 
        PP/TP  & \checkmark  &   $\circ$  & \checkmark  &   \XSolidBrush   \\ 
        ZeRO  & \checkmark  &   \checkmark &  \checkmark   &   \checkmark  \\
        All2All EP  & \checkmark  &  \XSolidBrush & $\circ$   &  \XSolidBrush  \\
        RingAttention  & \checkmark  &  \XSolidBrush & \XSolidBrush  &  \XSolidBrush  \\ 
        Ulysses  & \checkmark  &   \XSolidBrush &  \checkmark  &   \XSolidBrush  \\
        Fused Kernels  & \checkmark &  $\circ$  & $\circ$   & $\circ$    \\ 
        Stage Fusion  & \checkmark  &  \checkmark & \XSolidBrush  &  \XSolidBrush \\
        Partial Rollout  & \checkmark  &  \XSolidBrush & $\circ$  &  \XSolidBrush \\
        Load Balance  & \checkmark  &  \XSolidBrush &  \checkmark  &  \XSolidBrush \\
        Transfer Dock  & \checkmark &  \XSolidBrush & \XSolidBrush  &  \XSolidBrush \\
        Allgather-Swap  & \checkmark  &  \XSolidBrush & \XSolidBrush  &  \XSolidBrush \\
        \hline
        Dense Model  & \checkmark  &  \checkmark  & \checkmark   &  \checkmark  \\
        MoE Model  & \checkmark  &  $\circ$ & \checkmark   &  \XSolidBrush \\
        \Xhline{1pt}
        \end{tabular}
            \textbf{\caption{Comparison of the features of popular open-source RL systems, where MindSpeed RL is denoted MSRL and $\circ$ indicates that the feature is in development or partially supported.}}
\end{table}

\subsection{MindSpeed RL Framework}
The architecture of MindSpeed RL, which consists of four layers, is shown in Figure 6. The first layer is the Ascend NPU--based resource pool, and we use Ray to manage and allocate NPUs for distributed RL training. On the basis of the resource pool, MindSpeed RL constructs its generation engine and training engine with vLLM--Ascend and MindSpeed, respectively. As mentioned in the background section, we integrate numerous parallelization strategies, e.g., TP, PP, DP, CP, and EP, and fused kernels, e.g., FlashAttention, RMSNorm, RoPE, SwiGLU, MatmulAdd, and GMM, for computing efficiency. In addition, additional engines, e.g., SGLang and FSDP2, are being integrated for better usability. In the third layer, the dataflow is organized with multiple workers, including the resharding flow and sample flow. As mentioned above, we design the distributed transfer dock strategy and allgather--swap technique to address the dispatch overhead and redundant memory usage issues in this layer. Finally, various algorithm trainers are presented for users as the top layer of MindSpeed RL. \par
For clarity, MindSpeed RL and several popular open-source frameworks are compared in Table 2. As shown, we optimize MindSpeed RL systematically to increase the efficiency of computing, communication, and memory. 

\begin{figure}[htb]
\centering
\includegraphics[width=.45\textwidth, trim=0.7mm 0.7mm 0.7mm 0.7mm, clip]{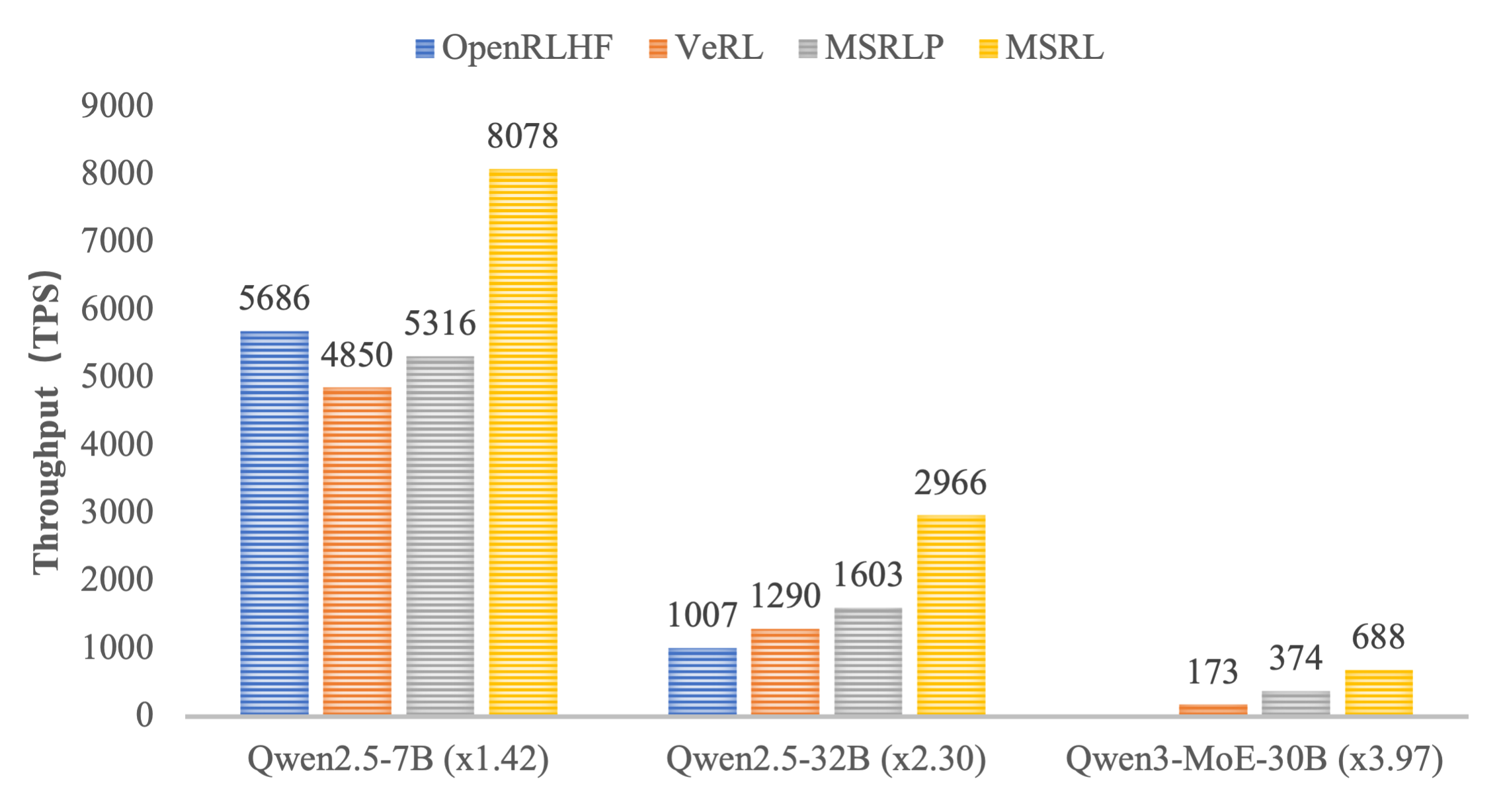}
\caption{End-to-end performance comparison for OpenRLHF, VeRL, MSRLP, and MSRL. Compared with the baselines, MSRL increases the throughput by $1.42 \sim 3.97$ times.}
\end{figure}

\begin{table}[!htb]
\centering
    \setlength{\tabcolsep}{2 mm}
    \centering
    \begin{tabular}{c|cc|cc}
        \Xhline{1pt}
        \multirow{2}*{Iteration} & \multicolumn{2}{c}{600}        &  \multicolumn{2}{c}{1000}  \\
                                             &  MSRL  & VeRL                                  & MSRL & VeRL     \\
         \hline
        MATH500 (Pass@1)  & 85.7  & 84.7 & 85.0  &  83.2  \\
        AIME24 (Avg@24) & 23.7  &  23.3 & 23.8  &  21.9 \\
        GPQA (Avg@4)& 47.4 &  45.9 & 48.1  &  47.6 \\
        \Xhline{1pt}
        \end{tabular}
            \textbf{\caption{Score comparison for MindSpeed RL and VeRL on three benchmarks.}}
\end{table}

\section{Evaluation}
Here, we first demonstrate the experimental setting and then present the end-to-end performance and ablation study of MindSpeed RL. Finally, we present the results of MindSpeed RL for large-scale MoE models.
\subsection{Experiment Setup}
We deploy RL systems on an Ascend NPU cluster for RL training with 48 nodes and 384 NPUs. Each node is equipped with 8 Ascend 128 GB NPUs. For practical measurements, the bandwidth of H2D and D2H is 50 GB/s, and the bandwidth between servers is 300 MB/s. Although MindSpeed RL supports various algorithms, we conduct performance and memory comparisons with the popular GRPO algorithm to avoid redundant results. In the experiments, we use a rule reward and DeepScaleR \cite{deepscaler2025} as the prompt dataset. In addition, the throughput $T$ (TPS) in this paper is defined as follows:
 \begin{equation}
 \begin{aligned}
T = G \times N \times (PL+SL) / ND / ETE,
\end{aligned} 
\end{equation} 
where $ND$ is the number of devices and where $ETE$ is the end-to-end time of one iteration. We average the $T$ values of five iterations as the final performance metric.

\begin{figure}[htb]
\centering
\includegraphics[width=.45\textwidth, trim=0.8mm 0.8mm 0.8mm 0.8mm, clip]{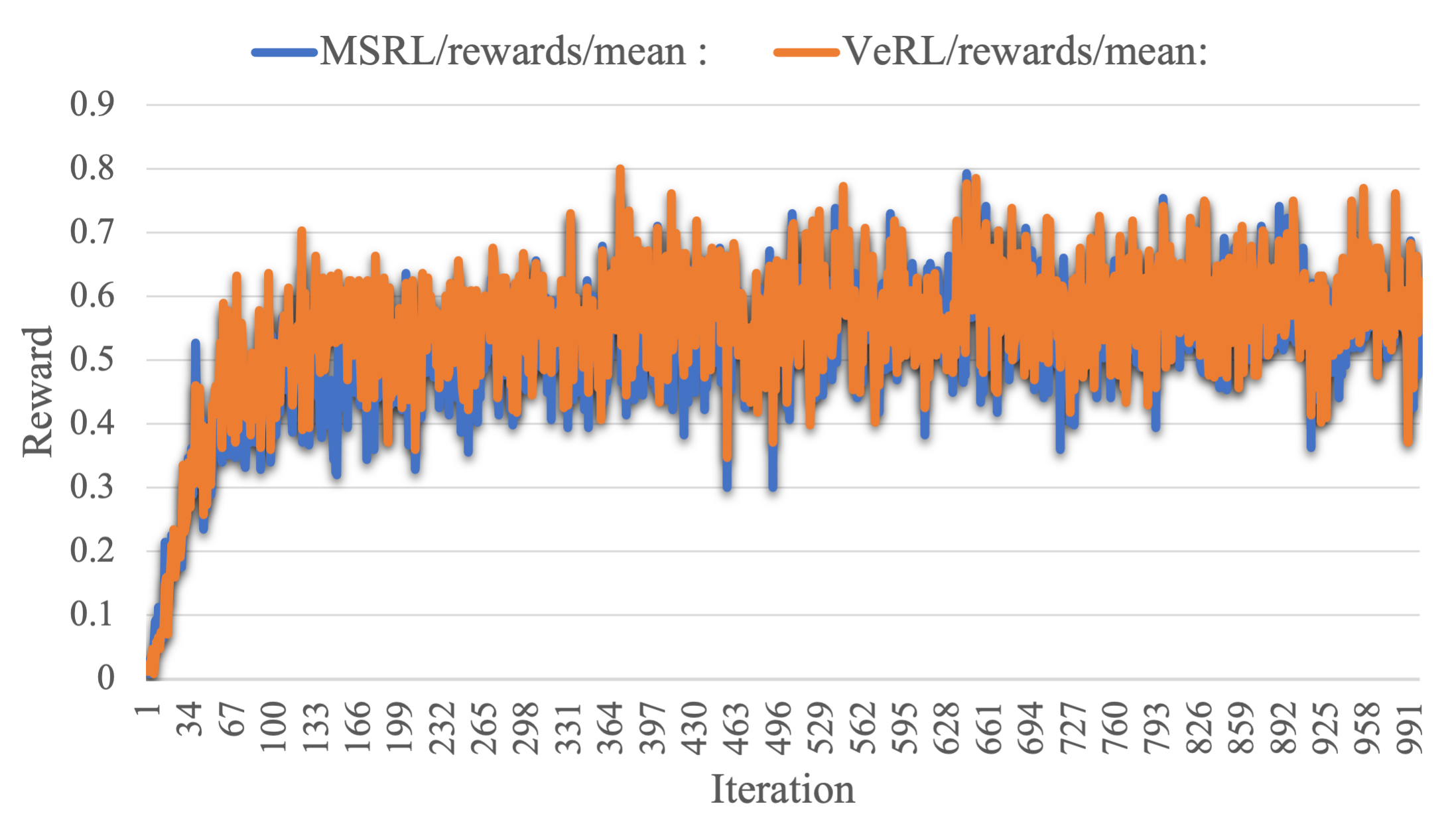}
\caption{Reward comparison for MindSpeed RL and VeRL with 1000 iterations.}
\end{figure}

\begin{figure}[htb]
\centering
\includegraphics[width=.48\textwidth, trim=0.5mm 0.5mm 0.5mm 0.5mm, clip]{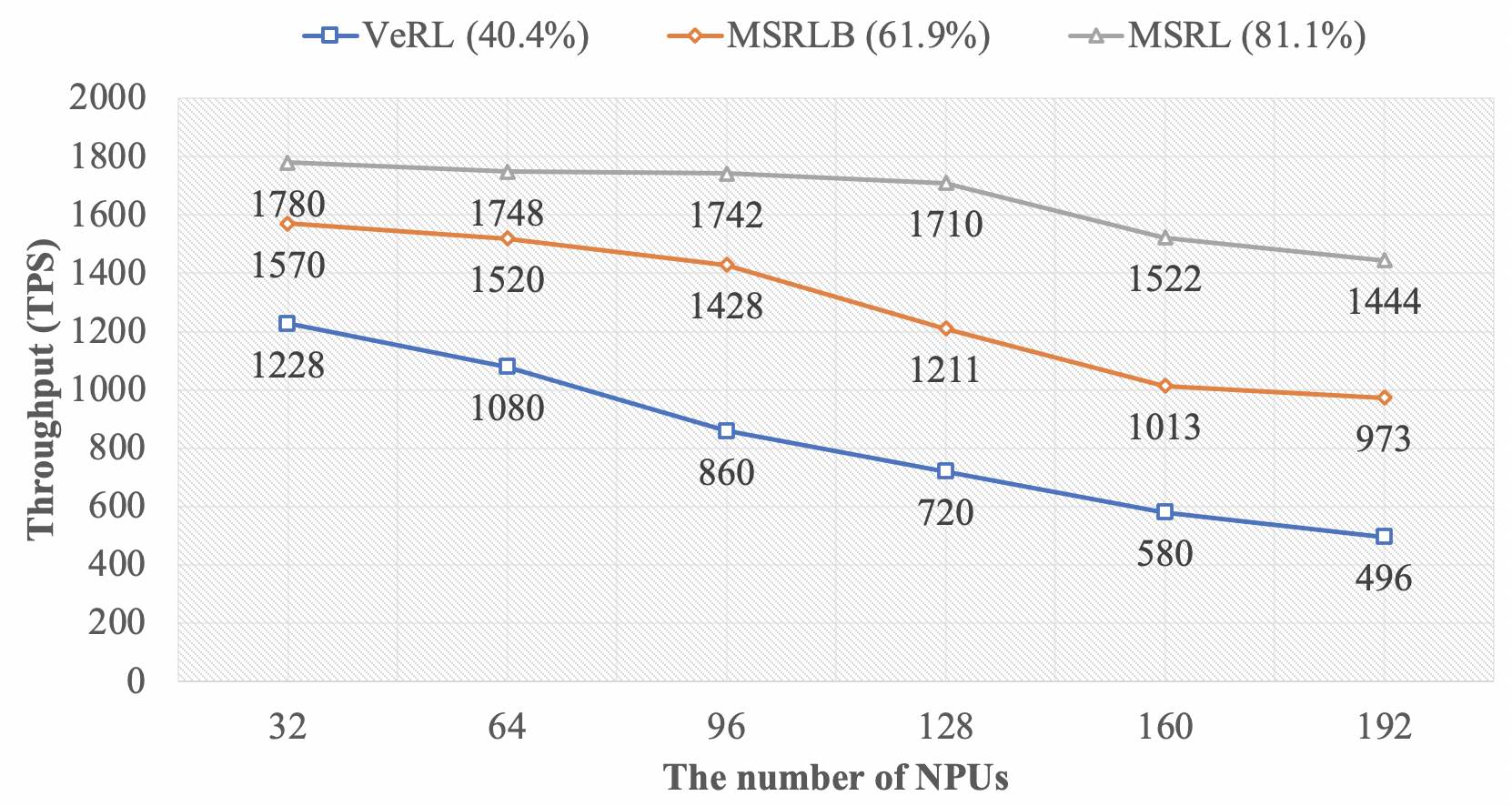}
\caption{Linearity comparison for VeRL, MSRLB, and MSRL.}
\end{figure}

\subsection{End-to-End Performance}
We compare MindSpeed RL (MSRL) with VeRL \cite{2024HybridFlow} and OpenRLHF \cite{2024OpenRLHF}. The performance of MindSpeed RL without the proposed two techniques (MSRLP) is also presented. Other open-source frameworks, such as DeepSpeed-Chat \cite{yao2023deepspeedchateasyfastaffordable}, NeMo-Aligner \cite{shen2024nemoalignerscalabletoolkitefficient}, TrlX \cite{trlx}, and Huggingface DDP \cite{wolf2020huggingfacestransformersstateoftheartnatural}, are not used here for comparison, since they are less representative and have been validated more slowly than the above baselines \cite{2024HybridFlow, yao2023deepspeedchateasyfastaffordable}. For each framework, we fine-tune the hyperparameters, e.g., parallelization strategies and fused kernels, to present the best performance on Qwen2.5-Dense-7B/32B and Qwen3-MoE-30B. In experiments, the global batch size $G$ is 256 and the number of responses for each prompt $N$ is 16.  The input length $PL$ and output length $SL$ are $2K$ and $8K$, respectively. The end-to-end performance when 16 NPUs are used are compared in Figure 7. As shown, on the basis of numerous acceleration techniques, MindSpeed RL generally increases the throughput of RL training by $1.42 \sim 3.97$ times, compared with the baselines. The results of MSRL and MSRLP show that the proposed transfer dock and allgather--swap methods contribute to the end-to-end performance by releasing dispatch overhead and increasing available memory for generation. \par

In addition, we compare the training processes and scores of MindSpeed RL and VeRL. The experiments are based on Qwen2.5-Dense-32B with $PL=2K$ and $SL=2K$. A comparison of the rewards of MindSpeed RL and VeRL is shown in Figure 8, and a comparison of 
scores on MATH 500 \cite{hendrycksmath2021}, AIME24 \cite{aime24}, and GPQA Diamond \cite{rein2023gpqagraduatelevelgoogleproofqa} is shown in Table 3. Compared with the baselines, MindSpeed RL has a stable training process and reliable scores for RL.

\begin{figure}[!htb]
\subfigure[Redundant memory in the resharding flow]{
\includegraphics[width=7.73cm]{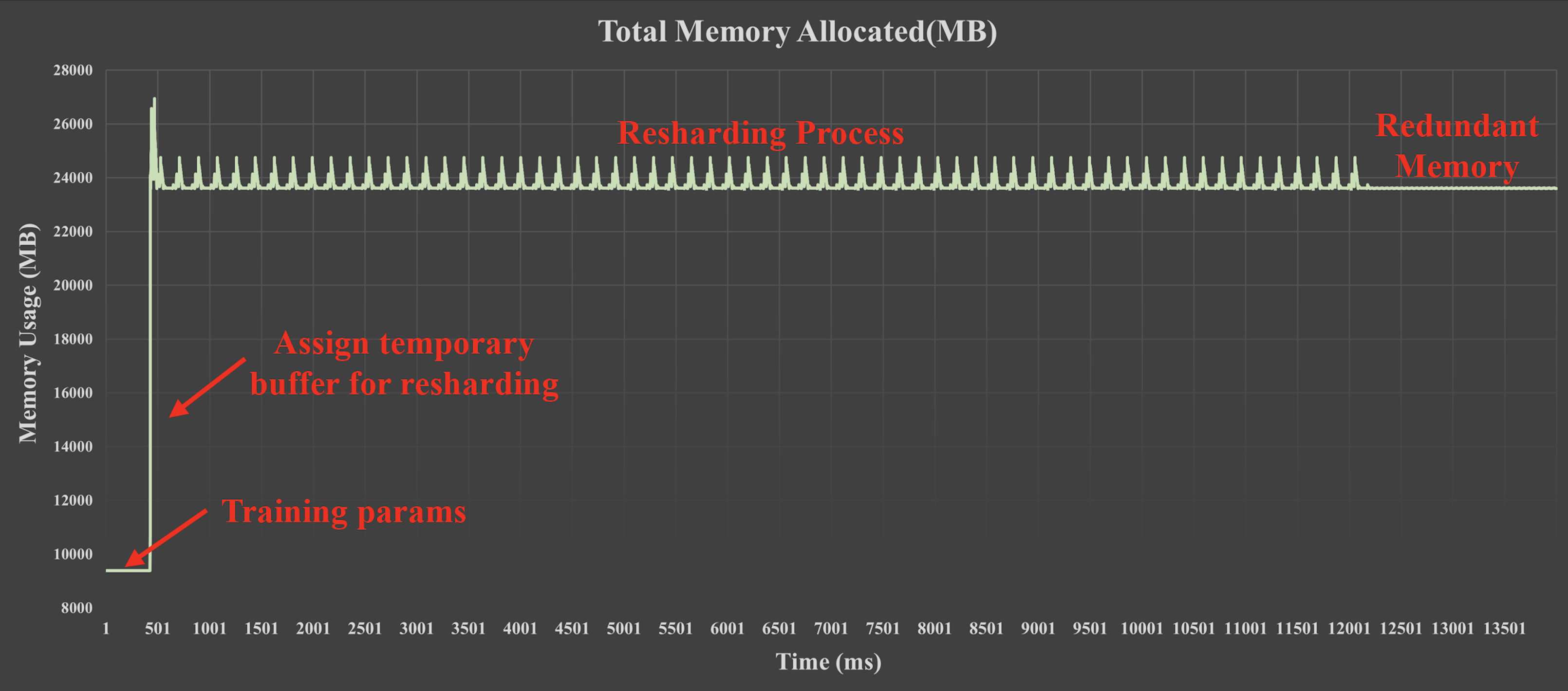}}
\subfigure[Released memory by the allgather-swap technique]{
\includegraphics[width=8cm]{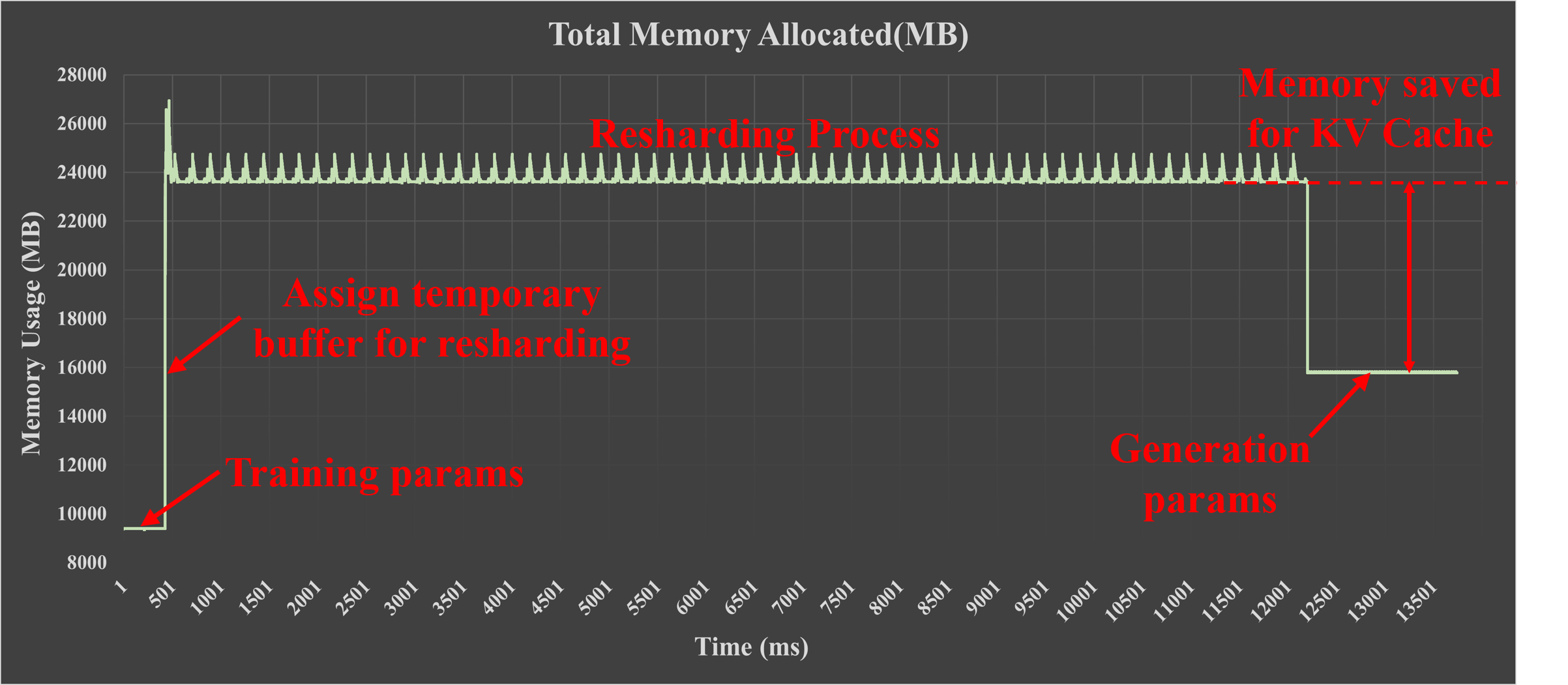}}
\caption{Memory profiling for the resharding flow of Qwen2.5-Dense-32B.}
\end{figure}

\subsection{Ablation Study}
For the transfer dock strategy, we show the linearity of MindSpeed RL, VeRL, and MindSpeed RL with the conventional replay buffer (MSRLB) in the cluster in Figure 9. In the experiments, we set the number of input prompts as 64 for each node and increase the number of nodes to test the linearity. As shown, the linearity of MindSpeed RL reaches 81.1\% with 192 NPUs, while those of VeRL and MSRLB are 40.4\% and 61.9\%, respectively. \par

For the allgather--swap technique, we compare the memory profiling in Figure 10.  The case is performed based on Qwen2.5-Dense-32B with a weight resharding from TP8DP2 to TP4DP4. As shown, using the proposed swap-technique, 8GB of redundant memory is released for the KV cache of each device in the resharding flow, which is essential for generation efficiency.
\begin{figure}[htb]
\centering
\includegraphics[width=.45\textwidth, trim=1mm 1mm 1mm 1mm, clip]{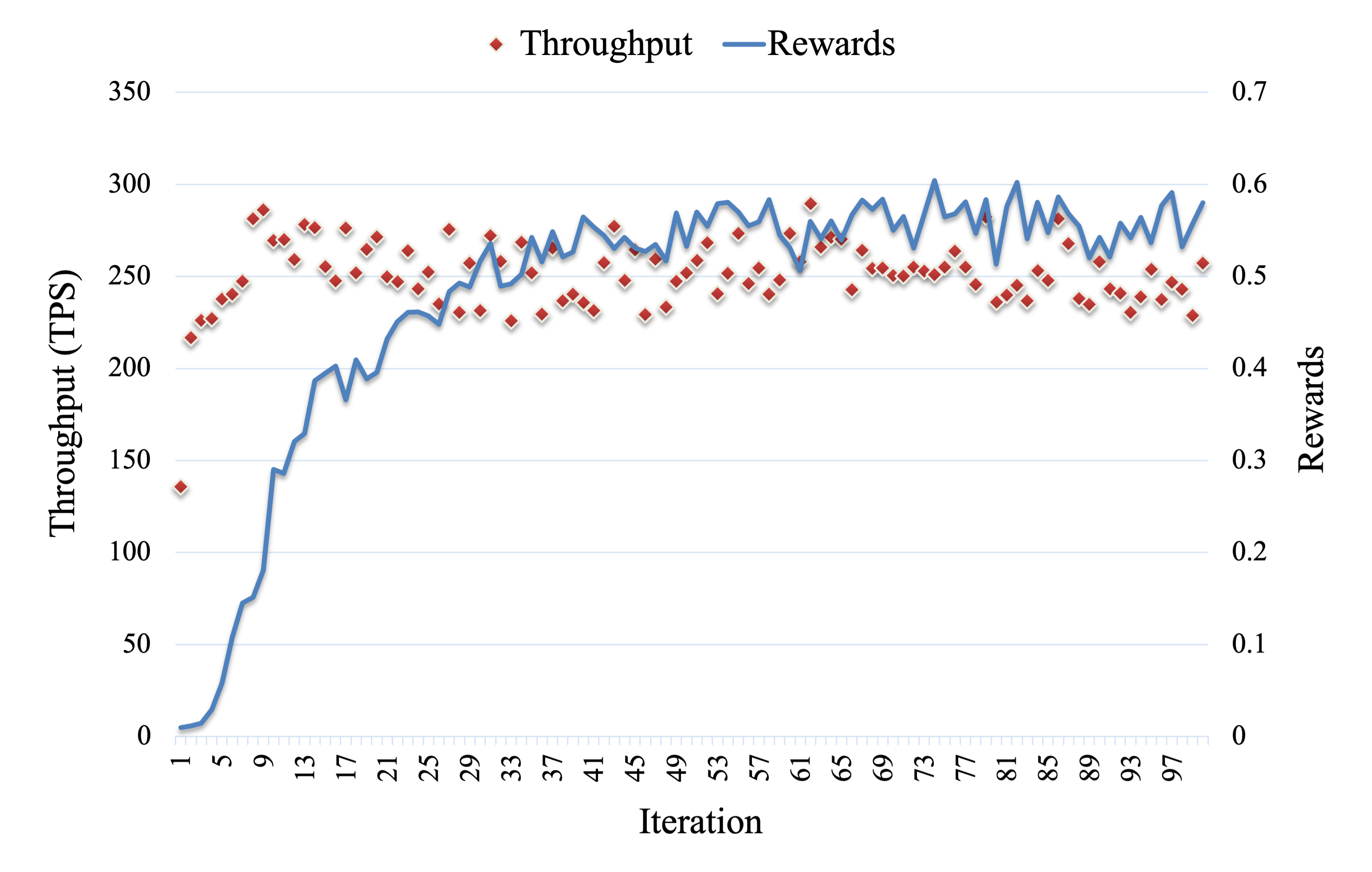}
\caption{Throughput and rewards of DeepSeek-R1-MoE-671B by MindSpeed RL with 384 NPUs.}
\end{figure}
\subsection{Results of Large-scale MoE Models}
Here, we show the performance of MindSpeed RL for a large-scale MoE model, i.e., DeepSeek-R1-MoE-671B, with 384 NPUs. In the experiments, the global batch size $G$ is 384 and the number of responses for each prompt $N$ is $32$. The input length $PL$ and output length $SL$ are $1K$ and $2K$, respectively. For the parallelization strategies, we use TP4PP6EP16DP2 in the update stage and TP2PP1EP64DP6 in the generation stage. The end-to-end throughput and rewards in 100 iterations are shown in Figure 10. As depicted, MindSpeed RL presents a steadily increasing reward curve for DeepSeek-R1-MoE-671B. The end-to-end throughput of MindSpeed RL fluctuates between 200 and 250 TPS, which is quite efficient. To the best of the author's knowledge, few open-source RL frameworks  support such a large MoE model.

\section{Conclusion}
This study introduces MindSpeed RL, a novel system for scalable and efficient RL training on Ascend NPU cluster. On the basis of the discussion on the dataflow bottlenecks, we improve the training performance of MindSpeed RL from three perspectives. First, a distributed transfer dock mechanism is designed to release the dispatch overhead in the sample flow. Second, an allgather-swap technique is proposed to eliminate the redundant memory usage in resharding flow. Finally, numerous acceleration techniques are integrated, including parallelization strategies and fused kernels. Compared with existing systems, large-scale experiments on popular models demonstrate the superior efficiency and scalability of MindSpeed RL. We open--source MindSpeed RL to encourage the development of novel RL algorithms on Ascend NPUs.

\bibliography{aaai2026.bib}

\begin{thebibliography}{53}
\providecommand{\natexlab}[1]{#1}

\bibitem[{{Ascend}(2024)}]{mindspeed}
{Ascend}. 2024.
\newblock MindSpeed.
\newblock \url{https://gitee.com/ascend/MindSpeed}.
\newblock Accessed: 2024-3-30.

\bibitem[{{Ascend}(2025)}]{vllmascend}
{Ascend}. 2025.
\newblock vLLM-Ascend.
\newblock \url{https://github.com/vllm-project/vllm-ascend/tree/main}.
\newblock Accessed: 2025-3-30.

\bibitem[{Dao(2023)}]{dao2023flashattention2fasterattentionbetter}
Dao, T. 2023.
\newblock FlashAttention-2: Faster Attention with Better Parallelism and Work
  Partitioning.
\newblock arXiv:2307.08691.

\bibitem[{DeepSeek-AI et~al.(2025)DeepSeek-AI, Guo, Yang, Zhang, Song, Zhang,
  Xu, Zhu, Ma, Wang, Bi, Zhang, and
  et~al.}]{deepseekai2025deepseekr1incentivizingreasoningcapability}
DeepSeek-AI; Guo, D.; Yang, D.; Zhang, H.; Song, J.; Zhang, R.; Xu, R.; Zhu,
  Q.; Ma, S.; Wang, P.; Bi, X.; Zhang, X.; and et~al., X.~Y. 2025.
\newblock DeepSeek-R1: Incentivizing Reasoning Capability in LLMs via
  Reinforcement Learning.
\newblock arXiv:2501.12948.

\bibitem[{Fu et~al.(2025)Fu, Gao, Shen, Zhu, Mei, He, Xu, Wei, Mei, Wang, Yang,
  Yuan, and Wu}]{fu2025areallargescaleasynchronousreinforcement}
Fu, W.; Gao, J.; Shen, X.; Zhu, C.; Mei, Z.; He, C.; Xu, S.; Wei, G.; Mei, J.;
  Wang, J.; Yang, T.; Yuan, B.; and Wu, Y. 2025.
\newblock AReaL: A Large-Scale Asynchronous Reinforcement Learning System for
  Language Reasoning.
\newblock arXiv:2505.24298.

\bibitem[{Graef, Wasielewski, and
  Clapp(2025)}]{graef2025flashnormfastnormalizationllms}
Graef, N.; Wasielewski, A.; and Clapp, M. 2025.
\newblock FlashNorm: Fast Normalization for LLMs.
\newblock arXiv:2407.09577.

\bibitem[{Havrilla et~al.(2023)Havrilla, Zhuravinskyi, Phung, Tiwari, Tow,
  Biderman, Anthony, and Castricato}]{trlx}
Havrilla, A.; Zhuravinskyi, M.; Phung, D.; Tiwari, A.; Tow, J.; Biderman, S.;
  Anthony, Q.; and Castricato, L. 2023.
\newblock TrlX: A Framework for Large Scale Reinforcement Learning from Human
  Feedback.
\newblock In \emph{Proceedings of the 2023 Conference on Empirical Methods in
  Natural Language Processing}, 8578--8595.

\bibitem[{Hendrycks et~al.(2021)Hendrycks, Burns, Kadavath, Arora, Basart, and
  Eric}]{hendrycksmath2021}
Hendrycks, D.; Burns, C.; Kadavath, S.; Arora, A.; Basart, S.; and Eric. 2021.
\newblock Measuring Mathematical Problem Solving With the MATH Dataset.
\newblock In \emph{Advances in Neural Information Processing Systems
  (NeurlPS)}.

\bibitem[{Hu et~al.(2025)Hu, Wu, Shen, Liu, Zhu, Wang, Jiang, Wang, Chen, Chen,
  Fang, Xianyu, Cao, Xu, and Liu}]{2024OpenRLHF}
Hu, J.; Wu, X.; Shen, W.; Liu, J.~K.; Zhu, Z.; Wang, W.; Jiang, S.; Wang, H.;
  Chen, H.; Chen, B.; Fang, W.; Xianyu; Cao, Y.; Xu, H.; and Liu, Y. 2025.
\newblock OpenRLHF: An Easy-to-use, Scalable and High-performance RLHF
  Framework.
\newblock arXiv:2405.11143.

\bibitem[{Huang et~al.(2019)Huang, Cheng, Bapna, Firat, Chen, Chen, Lee, Ngiam,
  Le, and Wu}]{2020GPipe}
Huang, Y.; Cheng, Y.; Bapna, A.; Firat, O.; Chen, M.~X.; Chen, D.; Lee, H.~J.;
  Ngiam, J.; Le, Q.~V.; and Wu, Y. 2019.
\newblock GPipe: Efficient Training of Giant Neural Networks using Pipeline
  Parallelism.
\newblock In \emph{Advances in Neural Information Processing Systems}, 8--14.
  Vancouver(CA).

\bibitem[{Jacobs et~al.(2024)Jacobs, Tanaka, Zhang, Zhang, Aminabadi, Song,
  Rajbhandari, and He}]{0System}
Jacobs, S.~A.; Tanaka, M.; Zhang, C.; Zhang, M.; Aminabadi, R.~Y.; Song, S.~L.;
  Rajbhandari, S.; and He, Y. 2024.
\newblock System Optimizations for Enabling Training of Extreme Long Sequence
  Transformer Models.
\newblock In \emph{2024 IEEE International Parallel and Distributed Processing
  Symposium Workshops (IPDPSW)}.

\bibitem[{Li et~al.(2023)Li, Fang, Bian, Liu, Liu, Huang, Wang, and
  You}]{2021Colossal}
Li, S.; Fang, J.; Bian, Z.; Liu, H.; Liu, Y.; Huang, H.; Wang, B.; and You, Y.
  2023.
\newblock Colossal-AI: A Unified Deep Learning System For Large-Scale Parallel
  Training.
\newblock In \emph{Proceedings of the 52nd International Conference on Parallel
  Processing}, 766--775.

\bibitem[{Liang et~al.(2021)Liang, Wu, Luo, Mika, Gonzalez, and
  Stoica}]{liang2021rllibflowdistributedreinforcement}
Liang, E.; Wu, Z.; Luo, M.; Mika, S.; Gonzalez, J.~E.; and Stoica, I. 2021.
\newblock RLlib Flow: Distributed Reinforcement Learning is a Dataflow Problem.
\newblock arXiv:2011.12719.

\bibitem[{Liu, Zaharia, and Abbeel(2023)}]{2023Ring}
Liu, H.; Zaharia, M.; and Abbeel, P. 2023.
\newblock Ring Attention with Blockwise Transformers for Near-Infinite Context.
\newblock arXiv:2310.01889.

\bibitem[{Luo et~al.(2025)Luo, Tan, Wong, Shi, Tang, Roongta, Cai, Luo, Li,
  Popa, and Stoica}]{deepscaler2025}
Luo, M.; Tan, S.; Wong, J.; Shi, X.; Tang, W.~Y.; Roongta, M.; Cai, C.; Luo,
  J.; Li, L.~E.; Popa, R.~A.; and Stoica, I. 2025.
\newblock DeepScaleR: Surpassing O1-Preview with a 1.5B Model by Scaling RL.
\newblock
  \url{https://pretty-radio-b75.notion.site/DeepScaleR-Surpassing-O1-Preview-with-a-1-5B-Model-by-Scaling-RL-19681902c1468005bed8ca303013a4e2}.
\newblock Notion Blog.

\bibitem[{{MATH-AI}(2024)}]{aime24}
{MATH-AI}. 2024.
\newblock AIME24.
\newblock \url{https://huggingface.co/datasets/math-ai/aime24}.

\bibitem[{Mei et~al.(2025)Mei, Fu, Li, Wang, Zhang, and
  Wu}]{mei2025realefficientrlhftraining}
Mei, Z.; Fu, W.; Li, K.; Wang, G.; Zhang, H.; and Wu, Y. 2025.
\newblock ReaL: Efficient RLHF Training of Large Language Models with Parameter
  Reallocation.
\newblock arXiv:2406.14088.

\bibitem[{Moritz et~al.(2018)Moritz, Nishihara, Wang, Tumanov, Liaw, Liang,
  Elibol, Yang, Paul, Jordan, and Stoica}]{Philipp2017Ray}
Moritz, P.; Nishihara, R.; Wang, S.; Tumanov, A.; Liaw, R.; Liang, E.; Elibol,
  M.; Yang, Z.; Paul, W.; Jordan, M.~I.; and Stoica, I. 2018.
\newblock Ray: A Distributed Framework for Emerging AI Applications.
\newblock arXiv:1712.05889.

\bibitem[{Narayanan et~al.(2021)Narayanan, Shoeybi, Casper, Legresley, and
  Zaharia}]{2021Efficient}
Narayanan, D.; Shoeybi, M.; Casper, J.; Legresley, P.; and Zaharia, M. 2021.
\newblock Efficient Large-Scale Language Model Training on GPU Clusters using
  megatron-lm.
\newblock In \emph{Proceedings of the International Conference for High
  Performance Computing, Networking, Storage and Analysis}.

\bibitem[{Noukhovitch et~al.(2025)Noukhovitch, Huang, Xhonneux, Hosseini,
  Agarwal, and Courville}]{noukhovitch2025asynchronousrlhffasterefficient}
Noukhovitch, M.; Huang, S.; Xhonneux, S.; Hosseini, A.; Agarwal, R.; and
  Courville, A. 2025.
\newblock Asynchronous RLHF: Faster and More Efficient Off-Policy RL for
  Language Models.
\newblock In \emph{Proceedings of the 2025 International Conference on Learning
  Representations (ICLR)}.

\bibitem[{{OpenAI}(2024)}]{openaio1}
{OpenAI}. 2024.
\newblock OpenAI o1.
\newblock \url{https://openai.com/zh-Hans-CN/o1/}.
\newblock Accessed: 2024-9-12.

\bibitem[{OpenAI et~al.(2024)OpenAI, Achiam, Adler, Agarwal, Ahmad, Akkaya,
  Aleman, Almeida, Altenschmidt, Altman, Anadkat, Avila, Babuschkin, Balaji,
  Balcom, Baltescu, Bao, Bavarian, Belgum, Bello, Berdine, Bernadett-Shapiro,
  Berner, Bogdonoff, Boiko, Boyd, Brakman, Brockman, Brooks, Brundage, Button,
  Cai, Campbell, Cann, Carey, Carlson, and
  et~al.}]{openai2024gpt4technicalreport}
OpenAI; Achiam, J.; Adler, S.; Agarwal, S.; Ahmad, L.; Akkaya, I.; Aleman,
  F.~L.; Almeida, D.; Altenschmidt, J.; Altman, S.; Anadkat, S.; Avila, R.;
  Babuschkin, I.; Balaji, S.; Balcom, V.; Baltescu, P.; Bao, H.; Bavarian, M.;
  Belgum, J.; Bello, I.; Berdine, J.; Bernadett-Shapiro, G.; Berner, C.;
  Bogdonoff, L.; Boiko, O.; Boyd, M.; Brakman, A.-L.; Brockman, G.; Brooks, T.;
  Brundage, M.; Button, K.; Cai, T.; Campbell, R.; Cann, A.; Carey, B.;
  Carlson, C.; and et~al. 2024.
\newblock GPT-4 Technical Report.
\newblock arXiv:2303.08774.

\bibitem[{Ouyang et~al.(2022)Ouyang, Wu, Jiang, Almeida, Wainwright, Mishkin,
  Zhang, Agarwal, Slama, Ray, Schulman, Hilton, Kelton, Miller, Simens, Askell,
  Welinder, Christiano, Leike, and Lowe}]{2022Training}
Ouyang, L.; Wu, J.; Jiang, X.; Almeida, D.; Wainwright, C.~L.; Mishkin, P.;
  Zhang, C.; Agarwal, S.; Slama, K.; Ray, A.; Schulman, J.; Hilton, J.; Kelton,
  F.; Miller, L.; Simens, M.; Askell, A.; Welinder, P.; Christiano, P.; Leike,
  J.; and Lowe, R. 2022.
\newblock Training language models to follow instructions with human feedback.
\newblock arXiv:2203.02155.

\bibitem[{Qi et~al.(2023)Qi, Wan, Huang, and Lin}]{2023Zero}
Qi, P.; Wan, X.; Huang, G.; and Lin, M. 2023.
\newblock Zero Bubble Pipeline Parallelism.
\newblock arXiv:2401.10241.

\bibitem[{Rajbhandari et~al.(2022)Rajbhandari, Li, Yao, Zhang, Aminabadi, Awan,
  Rasley, and
  He}]{rajbhandari2022deepspeedmoeadvancingmixtureofexpertsinference}
Rajbhandari, S.; Li, C.; Yao, Z.; Zhang, M.; Aminabadi, R.~Y.; Awan, A.~A.;
  Rasley, J.; and He, Y. 2022.
\newblock DeepSpeed-MoE: Advancing Mixture-of-Experts Inference and Training to
  Power Next-Generation AI Scale.
\newblock arXiv:2201.05596.

\bibitem[{Rajbhandari et~al.(2020)Rajbhandari, Rasley, Ruwase, and
  He}]{2020ZeRO}
Rajbhandari, S.; Rasley, J.; Ruwase, O.; and He, Y. 2020.
\newblock ZeRO: Memory Optimizations Toward Training Trillion Parameter Models.
\newblock In \emph{SC20: International Conference for High Performance
  Computing, Networking, Storage and Analysis}.

\bibitem[{Rasley et~al.(2020)Rasley, Rajbhandari, Ruwase, and
  He}]{2020DeepSpeed}
Rasley, J.; Rajbhandari, S.; Ruwase, O.; and He, Y. 2020.
\newblock DeepSpeed: System Optimizations Enable Training Deep Learning Models
  with Over 100 Billion Parameters.
\newblock In \emph{KDD 20: The 26th ACM SIGKDD Conference on Knowledge
  Discovery and Data Mining}.

\bibitem[{Rein et~al.(2023)Rein, Hou, Stickland, Petty, Pang, Dirani, Michael,
  and Bowman}]{rein2023gpqagraduatelevelgoogleproofqa}
Rein, D.; Hou, B.~L.; Stickland, A.~C.; Petty, J.; Pang, R.~Y.; Dirani, J.;
  Michael, J.; and Bowman, S.~R. 2023.
\newblock GPQA: A Graduate-Level Google-Proof QA Benchmark.
\newblock arXiv:2311.12022.

\bibitem[{Schulman et~al.(2017)Schulman, Wolski, Dhariwal, Radford, and
  Klimov}]{2017Proximal}
Schulman, J.; Wolski, F.; Dhariwal, P.; Radford, A.; and Klimov, O. 2017.
\newblock Proximal Policy Optimization Algorithms.

\bibitem[{Shao et~al.(2024)Shao, Wang, Zhu, and
  et~al.}]{shao2024deepseekmathpushinglimitsmathematical}
Shao, Z.; Wang, P.; Zhu, Q.; and et~al. 2024.
\newblock DeepSeekMath: Pushing the Limits of Mathematical Reasoning in Open
  Language Models.
\newblock arXiv:2402.03300.

\bibitem[{Shazeer(2020)}]{shazeer2020gluvariantsimprovetransformer}
Shazeer, N. 2020.
\newblock GLU Variants Improve Transformer.
\newblock arXiv:2002.05202.

\bibitem[{Shen et~al.(2024)Shen, Wang, Delalleau, Zeng, Dong, Egert, Sun,
  Zhang, Jain, Taghibakhshi, Ausin, Aithal, and
  Kuchaiev}]{shen2024nemoalignerscalabletoolkitefficient}
Shen, G.; Wang, Z.; Delalleau, O.; Zeng, J.; Dong, Y.; Egert, D.; Sun, S.;
  Zhang, J.; Jain, S.; Taghibakhshi, A.; Ausin, M.~S.; Aithal, A.; and
  Kuchaiev, O. 2024.
\newblock NeMo-Aligner: Scalable Toolkit for Efficient Model Alignment.
\newblock arXiv:2405.01481.

\bibitem[{Shen et~al.(2025)Shen, Liu, Wu, Zhu, Yang, Xin, Yue, and
  Yan}]{shen2025exploringdatascalingtrends}
Shen, W.; Liu, G.; Wu, Z.; Zhu, R.; Yang, Q.; Xin, C.; Yue, Y.; and Yan, L.
  2025.
\newblock Exploring Data Scaling Trends and Effects in Reinforcement Learning
  from Human Feedback.
\newblock arXiv:2503.22230.

\bibitem[{Sheng et~al.(2025)Sheng, Zhang, Ye, Wu, Zhang, Zhang, Peng, Lin, and
  Wu}]{2024HybridFlow}
Sheng, G.; Zhang, C.; Ye, Z.; Wu, X.; Zhang, W.; Zhang, R.; Peng, Y.; Lin, H.;
  and Wu, C. 2025.
\newblock HybridFlow: A Flexible and Efficient RLHF Framework.
\newblock In \emph{Proceedings of the Twentieth European Conference on Computer
  Systems}, EuroSys ’25, 1279–1297. ACM.

\bibitem[{Shoeybi et~al.(2019)Shoeybi, Patwary, Puri, Legresley, Casper, and
  Catanzaro}]{2019Megatron}
Shoeybi, M.; Patwary, M.; Puri, R.; Legresley, P.; Casper, J.; and Catanzaro,
  B. 2019.
\newblock Megatron-LM: Training Multi-Billion Parameter Language Models Using
  GPU Model Parallelism.
\newblock In \emph{Proceedings of CoRR}.

\bibitem[{Singh et~al.(2025)Singh, Khalid, Oyallon, and
  Belilovsky}]{singh2025modelparallelismsubnetworkdata}
Singh, V.; Khalid, Z.; Oyallon, E.; and Belilovsky, E. 2025.
\newblock Model Parallelism With Subnetwork Data Parallelism.
\newblock arXiv:2507.09029.

\bibitem[{Su et~al.(2023)Su, Lu, Pan, Murtadha, Wen, and
  Liu}]{su2023roformerenhancedtransformerrotary}
Su, J.; Lu, Y.; Pan, S.; Murtadha, A.; Wen, B.; and Liu, Y. 2023.
\newblock RoFormer: Enhanced Transformer with Rotary Position Embedding.
\newblock arXiv:2104.09864.

\bibitem[{Team et~al.(2025)Team, Du, Gao, Xing, Jiang, Chen, Li, Xiao, Du, and
  Liao}]{kimiteam2025kimik15scalingreinforcement}
Team, K.; Du, A.; Gao, B.; Xing, B.; Jiang, C.; Chen, C.; Li, C.; Xiao, C.; Du,
  C.; and Liao, C. 2025.
\newblock Kimi k1.5: Scaling Reinforcement Learning with LLMs.
\newblock arXiv:2501.12599.

\bibitem[{Wang et~al.(2025)Wang, Zhang, Zhang, Hu, Li, Zhang, Li, Wu, Wang, and
  Hovy}]{wang2025reinforcementlearningenhancedllms}
Wang, S.; Zhang, S.; Zhang, J.; Hu, R.; Li, X.; Zhang, T.; Li, J.; Wu, F.;
  Wang, G.; and Hovy, E. 2025.
\newblock Reinforcement Learning Enhanced LLMs: A Survey.
\newblock arXiv:2412.10400.

\bibitem[{Wei et~al.(2022)Wei, Wang, Schuurmans, Bosma, Chi, Le, and
  Zhou}]{2022Chain}
Wei, J.; Wang, X.; Schuurmans, D.; Bosma, M.; Chi, E.; Le, Q.; and Zhou, D.
  2022.
\newblock Chain of Thought Prompting Elicits Reasoning in Large Language
  Models.
\newblock In \emph{Advances in Neural Information Processing Systems,
  35:24824-24837, 2022}.

\bibitem[{Wolf et~al.(2020)Wolf, Debut, Sanh, Chaumond, Delangue, Moi, Cistac,
  Rault, Louf, Funtowicz, Davison, Shleifer, von Platen, Ma, Jernite, Plu, Xu,
  Scao, Gugger, Drame, Lhoest, and
  Rush}]{wolf2020huggingfacestransformersstateoftheartnatural}
Wolf, T.; Debut, L.; Sanh, V.; Chaumond, J.; Delangue, C.; Moi, A.; Cistac, P.;
  Rault, T.; Louf, R.; Funtowicz, M.; Davison, J.; Shleifer, S.; von Platen,
  P.; Ma, C.; Jernite, Y.; Plu, J.; Xu, C.; Scao, T.~L.; Gugger, S.; Drame, M.;
  Lhoest, Q.; and Rush, A.~M. 2020.
\newblock HuggingFace's Transformers: State-of-the-art Natural Language
  Processing.
\newblock arXiv:1910.03771.

\bibitem[{Yang et~al.(2025)Yang, Li, Yang, and
  et~al.}]{yang2025qwen3technicalreport}
Yang, A.; Li, A.; Yang, B.; and et~al. 2025.
\newblock Qwen3 Technical Report.
\newblock arXiv:2505.09388.

\bibitem[{Yao et~al.(2023)Yao, Aminabadi, Ruwase, Rajbhandari, Wu, Awan,
  Rasley, Zhang, Li, and Holmes}]{yao2023deepspeedchateasyfastaffordable}
Yao, Z.; Aminabadi, R.~Y.; Ruwase, O.; Rajbhandari, S.; Wu, X.; Awan, A.~A.;
  Rasley, J.; Zhang, M.; Li, C.; and Holmes, C. 2023.
\newblock DeepSpeed-Chat: Easy, Fast and Affordable RLHF Training of
  ChatGPT-like Models at All Scales.
\newblock arXiv:2308.01320.

\bibitem[{Yu et~al.(2025)Yu, Zhang, Zhu, Yuan, and
  et~al.}]{yu2025dapoopensourcellmreinforcement}
Yu, Q.; Zhang, Z.; Zhu, R.; Yuan, Y.; and et~al. 2025.
\newblock DAPO: An Open-Source LLM Reinforcement Learning System at Scale.
\newblock arXiv:2503.14476.

\bibitem[{Zhang et~al.(2025)Zhang, Du, Liu, Kwon, Mo, Wang, Liu, You, Li, and
  Long}]{zhang2025jengaeffectivememorymanagement}
Zhang, C.; Du, K.; Liu, S.; Kwon, W.; Mo, X.; Wang, Y.; Liu, X.; You, K.; Li,
  Z.; and Long, M. 2025.
\newblock Jenga: Effective Memory Management for Serving LLM with
  Heterogeneity.
\newblock arXiv:2503.18292.

\bibitem[{Zhang et~al.(2024)Zhang, Sheng, Liu, Li, Feng, Liu, Liu, Jia, Peng,
  Lin, and Wu}]{aframework}
Zhang, C.; Sheng, G.; Liu, S.; Li, J.; Feng, Z.; Liu, Z.; Liu, X.; Jia, X.;
  Peng, Y.; Lin, H.; and Wu, C. 2024.
\newblock A Framework for Training Large Language Models for Code Generation
  via Proximal Policy Optimization.
\newblock In \emph{Proceedings of NL2Code Workshop of ACM KDD 2024}.

\bibitem[{Zhang et~al.(2022)Zhang, Diao, Wu, Wang, and Lin}]{2022Accelerating}
Zhang, S.; Diao, L.; Wu, C.; Wang, S.; and Lin, W. 2022.
\newblock Accelerating Large-Scale Distributed Neural Network Training with
  SPMD Parallelism.
\newblock In \emph{Proceedings of the 13th Symposium on Cloud Computing}.

\bibitem[{Zhao et~al.(2024)Zhao, Gu, Varma, Luo, Huang, Xu, Wright,
  Shojanazeri, Ott, and Shleifer}]{fsdp2}
Zhao, Y.; Gu, A.; Varma, R.; Luo, L.; Huang, C.~C.; Xu, M.; Wright, L.;
  Shojanazeri, H.; Ott, M.; and Shleifer, S. 2024.
\newblock FSDP2.
\newblock
  \url{https://docs.pytorch.org/docs/stable/distributed.fsdp.fully_shard.html}.
\newblock Accessed: 2024-9-1.

\bibitem[{Zhao et~al.(2023)Zhao, Gu, Varma, Luo, Huang, Xu, Wright,
  Shojanazeri, Ott, Shleifer, Desmaison, Balioglu, Damania, Nguyen, Chauhan,
  Hao, Mathews, and Li}]{2023PyTorch}
Zhao, Y.; Gu, A.; Varma, R.; Luo, L.; Huang, C.-C.; Xu, M.; Wright, L.;
  Shojanazeri, H.; Ott, M.; Shleifer, S.; Desmaison, A.; Balioglu, C.; Damania,
  P.; Nguyen, B.; Chauhan, G.; Hao, Y.; Mathews, A.; and Li, S. 2023.
\newblock PyTorch FSDP: Experiences on Scaling Fully Sharded Data Parallel.
\newblock arXiv:2304.11277.

\bibitem[{Zheng et~al.(2024)Zheng, Yin, Xie, Sun, Huang, Yu, Cao, Kozyrakis,
  Stoica, Gonzalez, Barrett, and Sheng}]{2023arXiv231207104Z}
Zheng, L.; Yin, L.; Xie, Z.; Sun, C.; Huang, J.; Yu, C.~H.; Cao, S.; Kozyrakis,
  C.; Stoica, I.; Gonzalez, J.~E.; Barrett, C.; and Sheng, Y. 2024.
\newblock SGLang: Efficient Execution of Structured Language Model Programs.
\newblock arXiv:2312.07104.

\bibitem[{Zhong et~al.(2025{\natexlab{a}})Zhong, Zhang, Song, Hu, Jin, Wu,
  Chen, Chen, Zhou, Wan, Zhou, Jiang, Zhu, and
  Jiang}]{zhong2025streamrlscalableheterogeneouselastic}
Zhong, Y.; Zhang, Z.; Song, X.; Hu, H.; Jin, C.; Wu, B.; Chen, N.; Chen, Y.;
  Zhou, Y.; Wan, C.; Zhou, H.; Jiang, Y.; Zhu, Y.; and Jiang, D.
  2025{\natexlab{a}}.
\newblock StreamRL: Scalable, Heterogeneous, and Elastic RL for LLMs with
  Disaggregated Stream Generation.
\newblock arXiv:2504.15930.

\bibitem[{Zhong et~al.(2025{\natexlab{b}})Zhong, Zhang, Wu, Liu, Chen, Wan, Hu,
  Xia, Ming, Zhu, and Jin}]{zhong2025optimizingrlhftraininglarge}
Zhong, Y.; Zhang, Z.; Wu, B.; Liu, S.; Chen, Y.; Wan, C.; Hu, H.; Xia, L.;
  Ming, R.; Zhu, Y.; and Jin, X. 2025{\natexlab{b}}.
\newblock Optimizing RLHF Training for Large Language Models with Stage Fusion.
\newblock arXiv:2409.13221.

\bibitem[{Zhou and Yang(2024)}]{tensorrt}
Zhou, Y.; and Yang, K. 2024.
\newblock Exploring TensorRT to Improve Real-Time Inference for Deep Learning.
\newblock In \emph{2022 IEEE 24th Int Conf on High Performance Computing
  (HPCC)}.

\end{thebibliography}

\end{document}